\documentstyle[tree-dvips,mlap,ranlp97]{article}

\newlength{\texteqlength}%
{\begin{equation}%
\begin{minipage}[c]{\texteqlength}\begin{center}}%
{\end{center}\end{minipage}\end{equation}}%

{$\left[\hspace{.5ex}\begin{tabular}{l@{\hspace{1ex}:\hspace{1ex}}l}}%
{\end{tabular}\hspace{.3ex}\right]$}%

\newlength{\cufindent}
{$\left(\hspace{-0.1em}\begin{tabular}{l@{\hspace{1ex}:\hspace{1ex}}l}}%
{\end{tabular}\hspace{-0.1em}\right)$}%

\newenvironment{cuflist}%
{\begin{math}\left[\begin{tabular}{l}}%
{\end{tabular}\right]\end{math}}%

\newcommand{\cufclause}[2]%
{#1 := \\%
\hspace*{\cufindent}#2.}%

\newcommand{\featdecl}[2]
  {\begin{tabular}{c}
     #1 \\
     \fbox{\begin{feats}#2\end{feats}}
   \end{tabular}}

\newenvironment{feats}{\begin{tabular}{l@{\hspace{1ex}:\hspace{1ex}}l}}%
{\end{tabular}}%

\newcounter{maxnode} 
\newcounter{nodea}
\newcounter{nodeb}
\newcounter{nodec}
\newcounter{noded}
\newcounter{nodee}
\newcounter{nodef}
\newcounter{nodeg}
\newcounter{nodeh}
\newcounter{nodei}
\newcounter{nodej}
\newcounter{numberofnodes}   
\newcommand{\newnodenames}{
  \addtocounter{nodea}{\thenumberofnodes}
  \addtocounter{nodeb}{\thenumberofnodes}
  \addtocounter{nodec}{\thenumberofnodes}
  \addtocounter{noded}{\thenumberofnodes}
  \addtocounter{nodee}{\thenumberofnodes}
  \addtocounter{nodef}{\thenumberofnodes}
  \addtocounter{nodeg}{\thenumberofnodes}
  \addtocounter{nodeh}{\thenumberofnodes}
  \addtocounter{nodei}{\thenumberofnodes}
}

\newcommand{\oldnodenames}{
  \addtocounter{nodea}{-\thenumberofnodes}
  \addtocounter{nodeb}{-\thenumberofnodes}
  \addtocounter{nodec}{-\thenumberofnodes}
  \addtocounter{noded}{-\thenumberofnodes}
  \addtocounter{nodee}{-\thenumberofnodes}
  \addtocounter{nodef}{-\thenumberofnodes}
  \addtocounter{nodeg}{-\thenumberofnodes}
  \addtocounter{nodeh}{-\thenumberofnodes}
  \addtocounter{nodei}{-\thenumberofnodes}
                           }

\newcommand{\unatree}[2]{
      \begin{array}{c}
         #1
         \\
         |
         \\
         #2
      \end{array}
   }

\newcommand{\verticalspace}{$\begin{array}{c}\\\end{array}\hspace*{-1em}$}

\newcommand{\vdtree}[2]{%
\newnodenames%
\begin{array}{c}
    \node
      {\thenodea}
      {$#1$}
    \\ 
    \\
    \node
      {\thenodeb}
      {\makebox[1em]{\verticalspace}}
    \node
      {\thenodec}
      {$#2$\verticalspace}
    \node
      {\thenoded}
      {\makebox[1em]{\verticalspace}}
\end{array}%
\nodeconnect[b]{\thenodea}[l]{\thenodeb}%
\nodeconnect[b]{\thenodea}[r]{\thenoded}%
\nodeconnect[l]{\thenodeb}[l]{\thenodec}%
\nodeconnect[r]{\thenodec}[r]{\thenoded}%
\oldnodenames%
}%

\newcommand{\vddtree}[1]{%
\newnodenames%
\begin{array}{c}
    \node
      {\thenodea}
      {}
    \\ 
    \\
    \node
      {\thenodeb}
      {\makebox[1em]{\verticalspace}}
    \node
      {\thenodec}
      {$#1$\verticalspace}
    \node
      {\thenoded}
      {\makebox[1em]{\verticalspace}}%
\end{array}%
\nodeconnect[t]{\thenodea}[l]{\thenodeb}%
\nodeconnect[t]{\thenodea}[r]{\thenoded}%
\nodeconnect[l]{\thenodeb}[l]{\thenodec}%
\nodeconnect[r]{\thenodec}[r]{\thenoded}%
\oldnodenames%
}

\newcommand{\vdrtree}[1]{%
\newnodenames%
\begin{array}{c}
    \node
      {\thenodea}
      {$#1$}
    \\ 
    \\
    \node
      {\thenodeb}
      {\makebox[2.5em]{{}\verticalspace}}
\end{array}%
\nodeconnect[b]{\thenodea}[l]{\thenodeb}%
\nodeconnect[b]{\thenodea}[r]{\thenodeb}%
\nodeconnect[l]{\thenodeb}[r]{\thenodeb}%
\oldnodenames%
}

\newlength{\codeindent} 

\newcommand{\linkrel}[2] {\begin{array}{c} \node{a}{$#1$} \\ \\
\node{b}{$#2$}
 \end{array}
\makedash{1pt} \nodeconnect{a}{b} \makedash{0pt}}

\newcommand{\treepair}[2]{\left\langle\!#1#2\!\right\rangle}

\newcommand{\stackedtrees}[2]{\!\!\begin{array}{c}#1\\[-0.8em]#2\end{array}\!\!}

\newcommand{\lexemi}[1]{w_{#1}}

\newcommand{\grami}[1]{G_{#1}}

\newcommand{\infrule}[3]
   {#1 \;\;\stackrel{\rname{#3}}{\Longrightarrow} \;\; #2}

\newcommand{\rname}[1]{\mbox{{\it (#1)\/}}}

\newcommand{\categ}{x}

\newcommand{\categi}[1]{x_{#1}}     

\newcommand{\semi}[1]{X_{#1}}     

\title{A generation algorithm for f-structure representations}
\author{ Toni Tuells \\
IULA , Universitat Pompeu Fabra
\\ La Rambla 30-32\\
Barcelona 08028, Catalonia (Spain) \\
{\tt tuells@upf.es}}

\begin{document}
\bibliographystyle{ranlp,fullname}
\maketitle
\thispagestyle{empty}

\begin{abstract}
This paper shows that previously reported generation algorithms run into problems when dealing with f-structure representations. A generation algorithm that is suitable for this type of representations is presented: the Semantic Kernel Generation (SKG) algorithm. The SKG method has the same processing strategy as the Semantic Head Driven generation (SHDG) algorithm and relies on the assumption that it is possible to compute the Semantic Kernel (SK) and non Semantic Kernel (Non-SK) information for each input structure. \\
\end{abstract}

\section{Introduction}
In this paper we take up the problem of (tacticaly) generating a string
from an f-structure representation; we will show that generation algorithms that have already been described in the literature are not directly applicable to this type of representations, 
and we will propose a new generation algorithm: 
the Semantic Kernel Generation (SKG) algorithm. We will also show that since the SKG generator is guided by the {\em semantic-head} and {\em syntactic-head} relations, it can be seen both as a variant of the {\em semantic head driven generation algorithm}) (SHDG) \cite{Shieb:90} and the {\em syntactic-head driven generation algorithm} (SynHDG) \cite{Koenig:syn}.
A former version of this work was originally reported in \cite{Nicpers:96}, which also describes alternative generation algorithms for f-structure representations.

\section{The Semantic Kernel Generation Algorithm}
In order to show why former generators fail to generate from f-structures and why SKG works, we will assume the grammar fragment and lexical entries (originally described in \cite{Nicpers:96}) which are shown in Figures~\ref{Lexical entry},~\ref{grammar fragment}.\footnote{ For the grammar fragment only the relevant semantic information is shown.} 
As for the grammar fragment, note that {\em rules 1a and 1b} introduce modifiers at sentence level, and {\em rule 3} introduces modifiers at vp level.{\em Rule 2} combines the subject with the {\em vp}.{\em Rule 4} deals with the complements of a vp.

\begin{figure}[htb]
\footnotesize
\begin{center}
\begin{avm}
\footnotesize
\[
\sc cat:        & \sf v \cr 
\sc lex:        & \sf generate \cr 
\sc subcat:     & \< NP$_{\@1}$, NP$_{\@2}$ \> \cr 
\sc sem:        & \[ \sc pred:  & \sf generate \cr 
                     \sc arg1:  & \@1 \cr 
                     \sc arg2:  & \@2 
                  \]
\]\end{avm}

\vspace{.3in}
\begin{avm}
\[
\sc cat:        & \sf n \cr 
\sc lex:        & \sf sentence \cr 
\sc sem:        & \[ \sc rel:  & \sf sentence\cr 
                  \]
\]
\end{avm}
\end{center}
\caption{Lexical entries for {\em sentence} and {\em program}}
\label{Lexical entry}
\end{figure}


\begin{figure*}
\footnotesize
\begin{flushleft}\small\avmoptions{center}\sisterskip=0ex\daughterskip=5ex
\begin{tabular}{llll}
(1a) & \begin{avm}\[cat: & s \\ sem: & SEM \\ sem:mod: & [M\|MODS]\]\end{avm} & $\longrightarrow$ &
 \begin{avm}\[cat: & adv \\ sem: M\]\end{avm}, \begin{avm}\[cat: & s \\ sem: & SEM \\ sem:mod: & MODS\]\end{avm} 
\\ \\
(1b) & \begin{avm}\[cat: & s \\ sem: & SEM\\ sem:mod: & [M\|MODS]\]\end{avm} & $\longrightarrow$ & \begin{avm}\[cat: & s \\ sem: & SEM \\ sem:mod: & MODS\]\end{avm}, \begin{avm}\[cat: & adv \\ sem: M\]\end{avm} 
\\ \\
(2) & \begin{avm}\[cat: & s \\ sem: & SEM\]\end{avm} & $\longrightarrow$ &
 \begin{avm}\[cat: & np \\ sem: & SEM\_SUBJ\]\end{avm}, \begin{avm}\[cat: vp \\sem: SEM \\subcat: & [\begin{avm}\[cat: & np \\ sem: & SEM\_SUBJ\]\end{avm}]\]\end{avm}
\\ \\
(3) & \begin{avm}\[cat: & vp \\ sem: & SEM \\ sem:mod: & [MOD\|MODS]\]\end{avm} & $\longrightarrow$ & \begin{avm}\[cat: & adv \\ sem: & MOD\]\end{avm}, \begin{avm}\[cat: & vp \\ sem: & SEM \\ sem:mod: & MODS\]\end{avm}
\\ \\
(4) & \begin{avm}\[cat: & vp \\ sem: & SEM \\ subcat: & [SUBJ\|REST]\]\end{avm} & $\longrightarrow$ &
\begin{avm}\[cat: & vp \\ sem: & SEM \\ subcat: & [SUBJ,\begin{avm}\[cat: & X \\ sem: & SEM_X]\]\end{avm}\|REST]\]\end{avm}, \begin{avm}\[cat: & X \\ sem: & SEM_X\]\end{avm}
\\ \\
(5) & \begin{avm}\[cat: & vp \\ sem: & SEM \\ subcat: & SUBCAT\]\end{avm} & $\longrightarrow$ & \begin{avm}\[cat: & v \\ sem: & SEM \\ subcat: & SUBCAT\]\end{avm}
\\ \\
(6) & \begin{avm}\[cat: & np \\ sem: & SEM\]\end{avm} & $\longrightarrow$ &
\begin{avm}\[cat: & det\]\end{avm}, \begin{avm}\[cat: & n2 \\ sem: & SEM\]\end{avm} 
\\ \\
(7) & \begin{avm}\[cat: & n2 \\ sem: & SEM\]\end{avm} & $\longrightarrow$ &
\begin{avm}\[cat: & n \\ sem: & SEM\]\end{avm} 
\\ \\
(8) & \begin{avm}\[cat: & n2 \\ sem: & SEM \\ sem:mod: & [MOD\|MODS]\]\end{avm} & $\longrightarrow$ &
\begin{avm}\[cat: adj \\ sem: & SEM\]\end{avm}, \begin{avm}\[cat: & n2 \\ sem: & SEM \\ sem:mod: & MODS\]\end{avm} 
\end{tabular}
\end{flushleft}
\caption{Grammar Fragment.Only semantic information is shown}
\label{grammar fragment}
\end{figure*}
The analysis of the sentence {\em The little prolog program generated the complex 
sentence quickly} is shown in Figure~\ref{Input semantics}.\footnote{The example is due to Nicolas Nicolov.} Note that input semantics
represents the deep predicate argument structure of sentences for the generator; 
modifiers are contained in set-valued feature ``mod''.

\begin{figure*} 
\begin{center}
\begin{avm}
\footnotesize
\[
\sc mod:        & \< \sf quick \> \cr 
\sc pred:       & \sf generate \cr 
\sc arg1:       & \[ \sc def: & + \cr 
                     \sc mod: & \sf \< little, prolog \> \cr 
                     \sc rel: & \sf program \cr 
                  \] \cr 
\sc arg2:       & \[ \sc def: & +\cr 
                     \sc mod: & \< \sf complex \> \cr 
                     \sc rel: & \sf sentence 
          \]
\]
\end{avm}
\end{center}
\caption{Input semantics}
\label{Input semantics}
\end{figure*}
First we look at the results we obtain after applying former generation methods on 
 f-structure representations and then we describe the
SKG algorithm.

For expository purposes we will use the graphical notation used in \cite{Koenig:syn} to describe the generation algorithms. Following \cite{Koenig:syn}: we will assume that the syntax-semantics-relation for a given grammar is stated by pairs of trees. The left tree shows a local syntactic dependency, whereas the right tree defines a local semantic dependency. We also assume that there is a one-to-one mapping from the nonterminal leaf nodes of the syntactic tree to the leaf nodes of the local semantic tree. Note that this is only a graphical notation for the {\em rule-to-rule hypothesis}, i.e., the fact that in the grammar each syntactic rule is related to a semantic analysis rule. An example is given below:
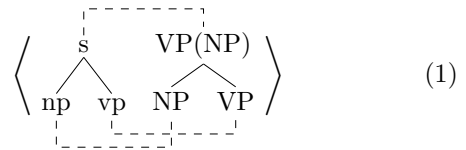
\begin{equation}
\begin{array}{c}
   \\
   \treepair
     {
  \begin{array}{cc}
   \multicolumn{2}{c}{
     \node
       {a}
       {s}}
   \\
   \\
    \node
      {b}
      {np}
    &
   \node
      {c}
      {vp}
  \end{array}
     }
     {
  \begin{array}{cc}
   \multicolumn{2}{c}{
     \node
       {d}
       {VP(NP)}}
   \\
   \\
    \node
      {e}
      {NP}
    &
   \node
      {f}
      {VP}
  \end{array}
     }
    \\ \mbox{}
\end{array}
  \nodeconnect[b]{a}[t]{b}
  \nodeconnect[b]{a}[t]{c}
  \nodeconnect[b]{d}[t]{e}
  \nodeconnect[b]{d}[t]{f}
  \makedash{3pt}
  \barnodeconnect[1em]{a}{d}
  \barnodeconnect[-1em]{b}{e}
  \barnodeconnect[-0.5em]{c}{f}
\end{equation}

The head-corner generator (\cite{Noord:th}, a variant of SHDG) and SynHDG are (graphically) described in Figure~\ref{hcgen} (taken directly from \cite{Koenig:syn}). The {\em lex} rule is the {\em prediction step} of the algorithm, i.e. it restricts the selection of lexical entries to those that can be linked with the local goal (visualized by a dotted line). The{\em hc\_complete} rule is the {\em bottom-up step} which selects a rule for which $\categ{h}$ is the syntactic head and $\semi{h}$ is the semantic head. As a result, it also predicts the head's sisters, which have to be expanded recursively ({\em top-down prediction}).
The difference between SHDG and SynHDG is the link relation for semantic structures: in \cite{Noord:th} the {\em semantic-based link relation} is defined as follows:
\begin{equation}
  link(\semi{},\semi{i})
  \mbox{\ \ {\em if\/}}
\end{equation}
$\semi{}$ and $\semi{i}$ are {\em identical}. If we represent semantics using first order terms, then we only have to check whether $\semi{}$ and $\semi{i}$ unify.
As for the SynHDG algorithm, the {\em semantic-based link relation} is defined as follows \cite{Koenig:syn}:
\begin{equation}
  link(\semi{},\semi{i})
  \mbox{\ \ {\em if\/}}
\end{equation}
$\semi{i}$ is a substructure of $\semi{}$. In practical terms, $\semi{i}$ is an element of the bag of semantic keywords that constitute $\semi{}$ \cite{Lex:95}.

\begin{figure*}
\footnotesize
all leaves are labeled with terminals and the tree does not contain
any dotted lines
\hspace{1em}
\rname{global-success} 
\\[1em]
\begin{math}
  \infrule
    {\treepair
       {\linkrel
          {\vddtree
             {\categ}
          }
          {\vdrtree
             {\categ}
          }
       }
       {\linkrel
          {\vddtree
             {\semi{}}
          }
          {\vdrtree
             {\semi{}}
          }
       }
    }
    {\treepair
        {\stackedtrees
           {\vddtree
              {\categ}}
           {\vdrtree
              {}}
       }
       {\stackedtrees
          {\vddtree
             {\semi{}}}
          {\vdrtree
             {}}
       }
    }
    {local-success} 
\end{math}
\\[.5em]
\begin{math}
\infrule
  {
   \treepair
     {
      \vddtree
        {\categ}
     }
     {
      \vddtree
        {\semi{}}
     }
  }
  {
   \treepair
     {\linkrel
        {\vddtree
           {\categ}
        }
        {
         \unatree
           {\categi{i}}
           {\lexemi{i}}
        }
     }
     {\linkrel
        {\vddtree
           {\semi{}}
        }
        {\semi{i}
        }
     }
   }
   {lex} 
   \hspace{1em}
   \mbox{{\em if\ }}
   \treepair
     {\unatree
        {\categi{i}}
        {\lexemi{i}}
     }
     {\semi{i}}
    \in \grami{} 
    \mbox{\ {\em and}\ } 
    link(\langle\categ,\semi{}\rangle,\langle\categi{i},\semi{i}\rangle)
\end{math}
\\[.5em]
\begin{math}
\infrule
  {
   \treepair
     {\linkrel
        {\vddtree{\categ}}
        {\vdrtree
           {\categi{h}}
        }
     }
     {\linkrel
        {\vddtree{\semi{}}}
        {\vdrtree
           {\semi{h}}
        }
     }
  }
  {
   \treepair
     {\linkrel
        {\vddtree{\categ}}
        {
         \begin{array}{c}
           \vdtree
             {\categi{0}}
             {\categi{1} \ldots \categi{h} \ldots \categi{n}} \\[-0.5em]
           \vdrtree
             {}
         \end{array}
        }
     }
     {\linkrel
        {\vddtree{\semi{}}}
        {\begin{array}{c}
           \vdtree
             {\semi{0}}
             {\semi{1} \ldots \semi{h} \ldots \semi{n}} \\[-0.5em]
           \vdrtree
             {}
         \end{array}
       }
     }
  }
  {hc\_complete}
\end{math}
\begin{flushright}
  {\em if\ }
   $\treepair
     {\vdtree
        {\categi{0}}
        {\categi{1} \ldots \categi{h} \ldots \categi{n}}}
     {\vdtree
        {\semi{0}}
        {\semi{1} \ldots \semi{h} \ldots \semi{n}}}
   \in \grami{}$
\end{flushright}
\caption{Head-Corner Generator
  ($\grami{}$ grammar description; 
   $\categi{i}$ syntactic category; 
   $\semi{i}$ semantic representation) (Taken from \cite{Koenig:syn}) }
\label{hcgen}
\end{figure*}

\subsection{The direct application of former generation procedures to
f-structure representations}
We will illustrate the problems SHDG (more specifically, a variant of it: the head-corner generator described in \cite{Noord:th}) runs into when dealing with f-structures by following its application to the input semantics given below (which corresponds to the np {\em the complex sentence}):
\begin{equation}
\label{input-complex}
\begin{avm}\[cat: & np \\ sem: & \begin{avm}\[rel: & sentence \\ def: & + \\ mod: & [complex]\]\end{avm}\]\end{avm}
\end{equation}
According to the algorithm described in Figure~\ref{hcgen}, and due to the {\em syntactic-head} and {\em semantic-head link relation}, the {\em lex} rule can only be applied to the lexical entry for {\em sentence} (Figure~\ref{Lexical entry}). However, since the semantic link relation is defined in terms of {\em unification}, applying rule {\em lex} leads to a new semantic goal which is {\em identical} to the input semantics. Next we need to apply to apply the {\em hc\_complete} rule; rule {\em 7} is the only possible candidate. After applying it, our current goal is the following:
\begin{equation}
\begin{avm}\[cat: & n2 \\ sem: & \begin{avm}\[rel: & sentence \\ def: & + \\ mod: & [complex]\]\end{avm}\]\end{avm}
\end{equation}
At this point, a new {\em hc\_complete} step needs to be taken. Now we have two candidates: rules {\em 6} and {\em 8}. If we select rule {\em 6}, and after generating recursively the determiner, we end up having generated only part of the sentence: {\em the sentence}. Rule {\em 8}, in its turn, can be {\em always} selected; consequently, we could end up having semantic goals that would look like that:
\begin{equation}
\begin{avm}\[cat: & n2 \\ sem: & \begin{avm}\[rel: & sentence \\ def: & + \\ mod: & [X,\ldots\|complex]\]\end{avm}\]\end{avm}
\end{equation}
In other words, {\em the generator would loop and would not terminate}.

The problems discussed above lead us to the following conclusion: the SHDG generator is {\em neither complete nor coherent}. These issues also arise with first order terms (see discussion in \cite{Shieb:90}); the problem here is that we lack the definition of grounded feature structures.

\subsection{Semantic Kernel Generator}

The main assumption behind the SKG algorithm is that the generator is capable of distinguishing 
between the following 
types of semantic information within input structures:
\begin{itemize}
\item {\bf Semantic Kernel (SK) Information}: Semantic structure completely
  which is predictible from the lexicon (i.e, there is at least one lexical
  entry which subsumes this structure).
\item {\bf Non Semantic Kernel (Non-SK) Information}: Semantic structure
  which is not predictible from any lexical entry (typically, lists).
  In our grammar, modifiers are represented as a list. This list is 
  Non-SK information.
\end{itemize}
Similarly, the generator is given the following information with respect to
the types of rules:
\begin{itemize}
\item {\bf SK Rules}: Rules which do not add Non-SK information.
\item {\bf Non-SK Rules}: Rules which add Non-SK information. Rules 1,3,8 in
 our grammar. 
\end{itemize}

The hypothesis behind this classification is that of {\em structural
  predictibility}: SK information comes from the lexicon (i.e, SK information can be seen as grounded feature terms), and non SK information is introduced through rules. In other words, the generator knows whether each type of input structure comes from a lexical entry or whether it has been constructed from a (non SK) rule. Thus, the restrictedness of the algorithm is due to the fact that it operates under the assumption
  that we can recursively decompose each input structure into SK and Non-SK
  information. 
  
A graphical version of the SKG algorithm is given in Figure~\ref{skgen}. 
\begin{figure*}
all leaves are labeled with terminals and the tree does not contain
any dotted lines
\hspace{2em}
\rname{global-success} 
\\[1em]
\begin{math}
  \infrule
    {\treepair
       {\linkrel
          {\vddtree
             {\categ}
          }
          {\vdrtree
             {\categ}
          }
       }
       {\linkrel
          {\vddtree
             {\semi{}}
          }
          {\vdrtree
             {\semi{}}
          }
       }
    }
    {\treepair
        {\stackedtrees
           {\vddtree
              {\categ}}
           {\vdrtree
              {}}
       }
       {\stackedtrees
          {\vddtree
             {\semi{}}}
          {\vdrtree
             {}}
       }
    }
    {local-success} 
\end{math}
\\[1em]
\begin{math}
\infrule
  {
   \treepair
     {
      \vddtree
        {\categ}
     }
     {
      \vddtree
        {\semi{}}
     }
  }
  {
   \treepair
     {\linkrel
        {\vddtree
           {\categ}
        }
        {
         \unatree
           {\categi{i}}
           {\lexemi{i}}
        }
     }
     {\linkrel
        {\vddtree
           {\semi{}}
        }
        {\semi{i}
        }
     }
   }
   {lex} 
   \\[1em]
   \mbox{{\em if\ }}
   \treepair
     {\unatree
        {\categi{i}}
        {\lexemi{i}}
     }
     {\semi{i}}
    \in \grami{} 
    \mbox{\ {\em and}\ } 
    link(\categ,\categi{i}) \mbox{\ {\em and}\ } sk(\semi{},\semi{i})
    \mbox{\ {\em and}\ } SK(\semi{})
\end{math}
\\[1em]
\begin{math}
\infrule
  {
   \treepair
     {\linkrel
        {\vddtree{\categ}}
        {\vdrtree
           {\categi{h}}
        }
     }
     {\linkrel
        {\vddtree{\semi{}}}
        {\vdrtree
           {\semi{h}}
        }
     }
  }
  {
   \treepair
     {\linkrel
        {\vddtree{\categ}}
        {
         \begin{array}{c}
           \vdtree
             {\categi{0}}
             {\categi{1} \ldots \categi{h} \ldots \categi{n}} \\[-0.5em]
           \vdrtree
             {}
         \end{array}
        }
     }
     {\linkrel
        {\vddtree{\semi{}}}
        {\begin{array}{c}
           \vdtree
             {\semi{0}}
             {\semi{1} \ldots \semi{h} \ldots \semi{n}} \\[-0.5em]
           \vdrtree
             {}
         \end{array}
       }
     }
  }
  {hc\_complete}
\end{math}
  \\[1em]
\begin{flushright}
  {\em if\ }
   $\treepair
     {\vdtree
        {\categi{0}}
        {\categi{1} \ldots \categi{h} \ldots \categi{n}}}
     {\vdtree
        {\semi{0}}
        {\semi{1} \ldots \semi{h} \ldots \semi{n}}}
   \in \grami{SK}$
    \mbox{\ {\em and}\ } 
    $SK(\semi{})$
\end{flushright}
\begin{math}
\infrule
  {
   \treepair
     {\linkrel
        {\vddtree{\categ}}
        {\vdrtree
           {\categi{h}}
        }
     }
     {\linkrel
        {\vddtree{\semi{}}}
        {\vdrtree
           {\semi{h}}
        }
     }
  }
  {
   \treepair
        {
         \begin{array}{c}
           \vdtree
             {\categi{}}
             {\categi{1} \ldots \categi{h} \ldots \categi{n}} \\[-0.5em]
           \vdrtree
             {}
         \end{array}
        }
        {\begin{array}{c}
           \vdtree
             {\semi{}}
             {\semi{1} \ldots \semi{h} \ldots \semi{n}} \\[-0.5em]
           \vdrtree
             {}
         \end{array}
       }
  }
  {hc\_complete}
\end{math}
  \\[1em]
\begin{flushright}
  {\em if\ }
   $\treepair
     {\vdtree
        {\categi{}}
        {\categi{1} \ldots \categi{h} \ldots \categi{n}}}
     {\vdtree
        {\semi{}}
        {\semi{1} \ldots \semi{h} \ldots \semi{n}}}
   \in \grami{}$
    \mbox{\ {\em and}\ } 
    $\neg$ $SK(\semi{})$
    \mbox{\ {\em and}\ }
   sk(\semi{},\semi{h}) 
\end{flushright}
\caption{Semantic Kernel Generator
  ($\grami{}$ grammar description; 
   $\categi{i}$ syntactic category; 
   $\semi{i}$ semantic representation;
   $SK(\semi{})$ is true if $\semi{}$ contains only SK information;
   $\grami{SK}$ grammar description with only SK rules; 
   $\grami{\neg SK}$ grammar description with only nonSK rules; 
   $sk(\semi{},\semi{h})$ is true if $\semi{h}$ is SK information of $\semi{}$
  (Adapted from \cite{Koenig:syn}) }
\label{skgen}
\end{figure*}
In sum, this is the information the generator needs to know about
  the grammar:
\begin{itemize}
\item link relation (head relation).
\item The SK and Non-SK substructures of a given semantic representation.
\item The distinction between SK rules and Non-SK rules.
\item The syntactic goals to generate SK and Non-SK information.
\item The syntactic goal we obtain after combining SK and Non-SK
  information.
\end{itemize}

In order to show how the SKG algorithm works we will follow its application to the input semantics for {\em the complex sentence} given in example ~\ref{input-complex}. For this input semantics, we have two SK structures:
\begin{equation}
\begin{avm}\[rel: & sentence \]\end{avm}
\end{equation}
\begin{equation}
\begin{avm}\[def: & + \]\end{avm}
\end{equation}
and one nonSK structure:
\begin{equation}
\begin{avm}\[mods: & [complex] \]\end{avm}
\end{equation}
The {\em lex} rule cannot be applied because of the SK structure condition: input semantics has nonSK information. Thus, we can only apply the second {\em hc\_corner step}. This forces us to start from {\em rule 6} and generate (top-down) the following goals:
\begin{equation}
\begin{avm}\[cat: & det \\ sem: & \begin{avm}\[def: & +\]\end{avm}\]\end{avm}
\end{equation}
\begin{equation}
\label{n2goal}
\begin{avm}\[cat: & n2 \\ sem: & \begin{avm}\[rel: & sentence \\ mods: & [complex]\]\end{avm}\]\end{avm}
\end{equation}
Note that the generator has been told about the relation between nonSK information and SK and nonSK rules, therefore it knows where the modifiers come from. The determiners generation is reduced to applying the {\em lex} rule. To generate the {\em n2} goal (example ~\ref{n2goal}) we proceed as before; this goal has nonSK information, so the generator starts from {\em rule 8} and generates (top-down) the appropriate subgoals:
\begin{equation}
\label{sentence-goal}
\begin{avm}\[cat: & n2 \\ sem: & \begin{avm}\[rel: & sentence\]\end{avm}\]\end{avm}
\end{equation}
\begin{equation}
\begin{avm}\[cat: & adj \\ sem: & \begin{avm}\[rel: & complex\]\end{avm}\]\end{avm}
\end{equation}
The generation of the subgoals above is straightforward, since they do not contain nonSK information and the {\em lex} rule can be applied without problems. The application of the {\em hc\_corner step} to each of the subgoals deserves further comments, since we are runing the risk of having {\em termination} problems. For example, once we have applied the {\em lex} rule and the {\em hc\_corner} step for {\em sentence} we obtain the goal in example ~\ref{sentence-goal}. One may wonder whether we could apply rule {\em 8} again and end up having subgoals like the following:
\begin{equation}
\begin{avm}\[cat: & n2 \\ sem: & \begin{avm}\[rel: & sentence \\ mods: & [X,\ldots]\]\end{avm}\]\end{avm}
\end{equation}
{\em This situation is not possible}, since if we only have SK information only SK rules can be applied, and rule {\em 8} is a nonSK rule (see conditions on {\em hc\_complete step (1)} in Figure~\ref{skgen}).

Another example will clarify how the generator works. Assuming we want to
generate a string for the semantic representation in
Figure~\ref{Input semantics},
the following sentences should be generated according to the grammar:\footnote{In this example we will only concentrate in the generation of {\em quickly} at sentence or vp level; the rest of modifiers ({\em complex, little, prolog}) for the np level  would be generated in an identical manner.} 
\begin{itemize}
\item {\em the \{little,prolog\} program generated the complex sentence quickly}.
\item {\em quickly the \{little,prolog\} program generated the complex sentence}.
\item {\em the \{little,prolog\} program quickly generated the complex sentence}.
\end{itemize}
The generator detects that the semantic representation consists of both
a SK structure and a Non-SK structure (``quickly''). Thus, according to
the grammar, there are several ways of generating these structures
given the original goal (which is a sentence). The generator tries the following combinations:
\begin{itemize}
\item It generates a ``S'' string for the SK information and a
  ``adv'' string for the Non-SK information. Both strings can
  be combined in two ways (which corresponds to rules R1a and R1b). This
  gives us two of the possibilities.
\item It generates a ``VP'' string for the SK information and
  ``adv'' string for the Non-SK information (this
  corresponds to rule R3). After generating these strings, and
  according to rule R2, we link the ``VP'' to ``S''
\end{itemize}

\section{Discussion}
The SKG generator proceeds top-down, generating the appropriate subgoals, when it finds nonSK information.
It proceeds bottom-up when lexical prediction can be made (when there is only SK information), and the {\em head-corner step} for SK information can only be performed using SK rules, thus avoiding {\em termination} problems.

The way our generator works and
the distinction between SK and Non-SK information resembles the definition
of the {\em restrictor} operator and the treatment of modifiers given in
 \cite{Wede:93}. The difference is obviously the context of application: In \cite{Wede:93}
the main interest is structural misalignment between f-structure and semantic
representations, whereas our concern is string generation from f-structure
representations.

\section{Implementation}
Our framework has been the Sicstus-Prolog version of the CUF language 
  \cite{Cuf:93} plus a layer on top of it which implements the grammar formalism, the (left-corner) parser and the SKG generator.

\section{Conclusions}
We have shown that for f-structure representations previously proposed 
generation algorithms run into problems, and therefore, we have presented an alternative:  
the SKG algorithm.
The main assumptions behind the algorithm is the following: for each semantic input it is possible to compute 
its SK and Non-SK information. We have also shown that our approach resembles
the definition of the {\em restrictor} operator and the treatment of modifiers given
in \cite{Wede:93} to deal with structural misaligments between f-structure
and semantic representations.

\section{Acknowledgments}
I would like to thank Esther K\"{o}nig, Michael Dorna and Nicolas Nicolov for valuable comments on the ideas presented in this paper, and to Cristina Corcoll for spellchecking it. Obviously, I am responsible for any mistake.

\end{document}